%% file: colm2026_conference.tex
\documentclass{article} % For LaTeX2e
\usepackage{aditi}
\usepackage[tt=false, type1=true]{libertine}   % Linux Libertine for body/sans
\usepackage{inconsolata}                        % Inconsolata for \texttt / code
\usepackage{tabularx}
\usepackage{booktabs}

\usepackage{microtype}
\usepackage{hyperref}
\usepackage{url}
\usepackage{subcaption}
\usepackage{graphicx}
\usepackage{booktabs}
\usepackage{wrapfig}
\usepackage[most]{tcolorbox}
\usepackage{lstlinebgrd} % <--- Add this line to define linebackgroundcolor

\tcbset{colback=blue!2,colframe=blue!40!black,boxrule=0.5pt,arc=1pt}
\newtcolorbox{defbox}[1][]{
    enhanced, breakable,
    colback=highlightred,
    colframe=themedred,
    boxrule=0.5pt, arc=1pt,
    fonttitle=\bfseries,
    title=#1
}
\newcommand{\methods}{NULLs}

\usepackage{listings}
\definecolor{boxbg}{RGB}{248,248,245}
\definecolor{themedred}{HTML}{A70F29}
\colorlet{highlightred}{themedred!12!white}     % true light tint of themedred
\definecolor{badgegray}{RGB}{215,215,210}
\definecolor{commentgray}{RGB}{135,135,135}
\definecolor{kwpurple}{RGB}{125,80,200}
\definecolor{strgreen}{RGB}{40,140,80}
\definecolor{sinkonblue}{HTML}{9FAECC}
\definecolor{sinkoffpink}{HTML}{DC85B3}
\lstdefinestyle{nullspy}{
    language=Python,
    basicstyle=\ttfamily\footnotesize,
    keywordstyle=\color{kwpurple}\bfseries,
    commentstyle=\color{commentgray}\itshape,
    stringstyle=\color{strgreen},
    showstringspaces=false,
    keepspaces=true,
    columns=fullflexible,
    aboveskip=0.4em,
    belowskip=0.4em,
}
\usepackage{lineno}

\definecolor{darkblue}{rgb}{0, 0, 0.5}
\hypersetup{colorlinks=true, citecolor=darkblue, linkcolor=darkblue, urlcolor=darkblue}

\title{Natively Unlearnable Large Language Models}

\begin{document}
\setauthors{Gaurav Ghosal$^{1}$ \authorsep Pratyush Maini$^{2}$ \authorsep Aditi Raghunathan$^1$}
\setaffils{$^{1}$ Carnegie Mellon University \affilsep $^{2}$ DatologyAI}
%\ifcolmsubmission
%\linenumbers
%\fi
\setemail{gghosal@andrew.cmu.edu}
\setcode{https://github.com/AR-FORUM/NULLS}
\let\ForumWebsite\empty % no website field (force-empty so the title-block globe icon is suppressed)
\maketitle

\input{sections/abstract}

\input{sections/intro_new}

\section{Related Works}
\label{sec:related}

\textbf{Post-hoc unlearning.}
 Post-hoc methods modify a fully trained model to remove targeted information after training. One approach is gradient-based fine-tuning with losses that encourage lower probability on the target \citep{zhang2024npo, jang2022knowledgeunlearningmitigatingprivacy, yao2024large, eldan2023whosharrypotterapproximate}. Another line of work aims to localize the unlearning target to specific parameters and remove or modify them selectively \citep{chang-etal-2024-localization, maini2023neuralnetworkmemorizationlocalized, meng2022locating}. Despite extensive research, these approaches exhibit two opposing failure modes. First, they often impact the model beyond their intended target, leading to the degradation of semantically related knowledge \citep{maini2024tofu} and general capabilities \citep{shi2024musemachineunlearningsixway}. Second, post-hoc methods have proven easy to reverse. \citet{patil2023sensitiveinformationdeletedllms} show that information can remain accessible in the intermediate layers of models. Likewise, \citet{fan2025llmunlearningresilientrelearning} find that post-hoc unlearning methods fail to be robust under further fine-tuning attacks. This fragility has been observed in benign settings: \citet{zhang2025catastrophic} demonstrate that quantization can recover ostensibly unlearned information.
 
 \textbf{Source isolation.} To address the limitations of post-hoc unlearning, an emerging line of work aims to localize information during model training. In mixture-of-experts models, \citet{shi2025flexolmo, gururangan2021demixlayersdisentanglingdomains} allocate separate expert modules to different data sources and domains. Similarly, in a dense model, \citet{cloud2024gradientrouting, shilov2025datafilteringknowledgelocalization} route data from specific sources to a subset of model parameters by masking training gradients. Both approaches make unlearning straightforward by simply deleting the corresponding model components. However, they are limited in the granularity they support, as each source requires an individualized expert or set of neurons. Moreover, these approaches eliminate joint learning across sources by completely isolating the parameters that different sources update. \methods{} allows joint learning through a pool of shared backbone neurons, and enables better scaling by localizing source-specific knowledge to sparse masks in a shared pool of sink neurons.

\section{Natively Unlearnable LLMs}
\label{sec:method}
\subsection{Problem Framing}
\label{subsec:framing}
\textbf{Pretraining Data and Sources} Let $\mathcal{D}$ denote the full pre-training dataset. We assume that the documents in $\mathcal{D}$ can be partitioned into a set of non-overlapping \emph{sources} $S_1, \dots, S_N$ such that $\mathcal{D} = \bigcup_{i=1}^{N} S_i$. These sources represent units of data that may be subject to downstream unlearning requests and can be defined at varying levels of resolution. For instance, sources may correspond to individual documents or topically coherent clusters of data. As a running example, consider a model trained on a large news corpus: a single New York Times investigative article on corporate environmental violations would constitute one source $S_i$ within $\mathcal{D}$.

\textbf{Unlearning} Given a model $\Theta$ trained on $\mathcal{D}$ and a forget source $S_{\textrm{forget}}$, unlearning aims to obtain a model that behaves as if $S_{\textrm{forget}}$ were not present in the training corpus. In our example, the New York Times may issue a takedown request, designating the article as $S_{\textrm{forget}}$. The unlearned model should no longer reproduce distinctive passages or recall details reported exclusively in the article, such as the names of internal whistleblowers or proprietary data. However, it should preserve general knowledge of environmental regulation and corporate compliance acquired from other sources in $\mathcal{D} \setminus S_{\textrm{forget}}$. The gold standard is retraining on $\mathcal{D} \setminus S_{\textrm{forget}}$ to produce $\Theta_{\textrm{retrain}}$, but this is typically infeasible. Instead, prior work performs an update $\mathcal{U}(\Theta, S_{\textrm{forget}})$ that approximates $\Theta_{\textrm{retrain}}$ without retraining, using either gradient-based tuning or parameter editing.

\textbf{Natively Unlearnable LLMs} Post-hoc unlearning methods  \citep{maini2023neuralnetworkmemorizationlocalized,chang-etal-2024-localization} often degrade broader model capabilities and knowledge. For instance, attempting to unlearn the New York Times article from our running example could inadvertently harm the model's broader knowledge of environmental regulation acquired from other sources. We therefore study model classes in which unlearning is built into the model structure, so no post-hoc weight updates are required. We refer to such models as \textbf{natively unlearnable models}.

\noindent Prior work has attempted to achieve native unlearnability by dedicating separate experts or parameter subsets to each source \citep{shi2025flexolmo, cloud2024gradientrouting}. While effective when sources are few and coarsely defined, this strategy is impractical at the scale and granularity of language-model pretraining. First, such an approach scales poorly, as the parameter count grows linearly in the number of sources, which can number in the millions. Second, isolating sources in this manner prevents the model from acquiring general capabilities that span the corpus, since no parameters are shared across sources. To be practical, a natively unlearnable model must \emph{simultaneously} learn general capabilities across sources while preserving independent control over individual sources.

\subsection{Implementing \methods{}}

\begin{figure}[t]
\centering
\begin{minipage}[t]{0.48\linewidth}
\begin{tcolorbox}[
    colback=boxbg, colframe=boxbg, boxrule=0pt, arc=2mm,
    title={\sffamily\colorbox{badgegray}{\strut~\bfseries\scriptsize BEFORE~}\quad\textbf{Standard LLaMA MLP}},
    coltitle=black, colbacktitle=boxbg, fonttitle=\sffamily,
]
\begin{lstlisting}[style=nullspy]
def forward(self, x):
    x_fc_1 = self.fc_1(x)
    x_fc_2 = self.fc_2(x)
    x = F.silu(x_fc_1) * x_fc_2

    return self.proj(x)
\end{lstlisting}
\end{tcolorbox}
\end{minipage}\hfill
\begin{minipage}[t]{0.48\linewidth}
\begin{tcolorbox}[
    colback=boxbg, colframe=boxbg, boxrule=0pt, arc=2mm,
    title={\sffamily\colorbox{themedred}{\strut~\color{white}\bfseries\scriptsize AFTER~}\quad\textbf{\methods{} MLP}},
    coltitle=black, colbacktitle=boxbg, fonttitle=\sffamily,
]
\begin{lstlisting}[
    style=nullspy,
    escapeinside={(*@}{@*)},
    linebackgroundcolor={\ifnum\value{lstnumber}=5\color{highlightred}\else\color{boxbg}\fi},
]
def forward(self, x, source_id):
    x_fc_1 = self.fc_1(x)
    x_fc_2 = self.fc_2(x)
    x = F.silu(x_fc_1) * x_fc_2
    (*@\textcolor{themedred}{\ttfamily\footnotesize\bfseries x\ =\ x\ *\ mask(source\_id)}@*)
    return self.proj(x)
\end{lstlisting}
\end{tcolorbox}
\end{minipage}
\caption{\textbf{\methods{} requires minimal architectural modifications.} \methods{} modifies only the fully connected layers of the transformer. The post-nonlinearity activations are multiplied by a source-dependent mask which activates all shared backbone 
neurons but only a consistent fraction of the sink neuron pool.  We create the mask with a pseudo-random number generator, allowing it to be generated on the fly during training or inference. All other components of the transformer architecture remain unmodified.}
\label{fig:null-minimal-intervention}
\end{figure}

We implement \methods{} based on the Memorization Sinks architecture introduced in \citet{ghosal2025memorizationsinksisolatingmemorization}. Their work showed that selectively activating a pool of sink neurons can isolate broad memorization from a shared backbone. However, sink activation in Memorization Sinks is not tied to data provenance: there is no mechanism to identify which sources contributed to which sink neurons. As a result, Memorization Sinks does not enable selective access to information learned from individual sources. \methods{} closes this gap by assigning each source a deterministic sparse mask to the sink pool, generated from the source identifier alone. This makes source-specific knowledge individually addressable and removable without modifying any weights.

\textbf{Architecture} We target neurons in the transformer fully connected layers (MLPs) to implement native unlearnability, leaving the remainder of the architecture unmodified. This design choice follows from existing findings that MLP layers serve as the site of knowledge and memorization in transformers \citep{nanda2023factfinding,geva-etal-2021-transformer}. We partition the MLP hidden neurons at each layer into two sets: a \textbf{shared backbone} of $N_{\textrm{gen}}$ neurons which seeks to aggregate general capabilities and a \textbf{memorization sink pool} of $N_{\textrm{pool}}$ neurons which is selectively activated to induce a correspondence between sources and subsets of neurons within it.

\textbf{Activation of Sinks} For each source, we activate a subset of size $N_{\textrm{source}}$ of the $N_{\textrm{pool}}$ sink neuron pool while dropping out the remainder. This mask is generated deterministically by using the source identifier as a seed to a pseudo-random number generator. The $N_{\textrm{gen}}$ shared backbone neurons remain active across all examples to allow learning general capabilities. We refer to the ratio $\frac{N_{\textrm{source}}}{N_{\textrm{pool}}}$ as the \textit{overlap ratio}: it controls the expected overlap fraction between masks for different sources. This selective activation links information from a source $S_i$ to a specific sink activation mask, generating an explicit and known localization within the model. The pseudo-random generation of the masks ensures that even semantically related sources receive independent masks and reduces unintended knowledge entanglement. For a fixed pool size ($N_{\textrm{pool}}$) and $N_{\textrm{source}}$, the number of possible masks is combinatorial, enabling \methods{} to scale to many distinct sources with independently controllable representations.

\textbf{Inference-time Activation and Unlearning} At inference time, source-specific information is accessed by applying the source's mask to the sink pool. As a result, unlearning can be implemented by ensuring that the mask corresponding to the target source is not applied during the forward pass, enabling \methods{} to perform unlearning without modifying any model weights. Source-specific information can also be removed from the model permanently by zeroing out parameters associated with neurons that are active in the source's mask. Throughout our experiments, we evaluate two inference modes: \textbf{Sink-On}, in which the ground-truth source sink is activated, and \textbf{Sink-Off}, in which the next-closest source (by embedding similarity) is activated instead.

\section{Experiment Results}
\label{sec:results}
We validate \methods{} across two case studies that simulate different unlearning use-cases. Our Wikipedia case study tests whether \methods{} enables surgical removal of fine-grained sources (6M individual Wikipedia articles) despite substantial semantic overlap between them. Our Harry Potter case study then tests whether \methods{} enables robust removal of larger, topically connected sets of data.

\subsection{Article-Level Unlearning in Wikipedia}
\label{subsec:wikires}

\begin{figure*}[t]
\centering
\subcaptionbox{Shared Facts\label{fig:wikihistogramshared}}{
\centering
\includegraphics[width=0.31\linewidth]{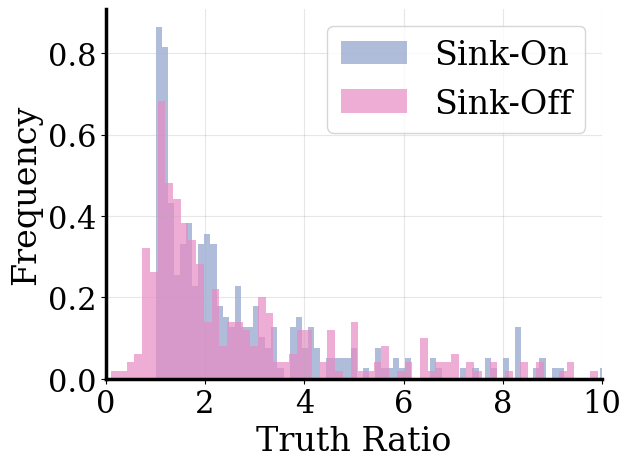}
}
\subcaptionbox{Article-Specific Facts\label{fig:wikihistogramisolated}}{
\centering
\includegraphics[width=0.31\linewidth]{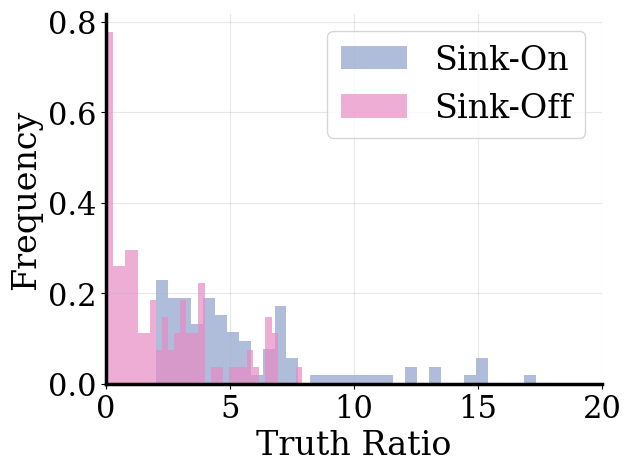}
}
\subcaptionbox{Gradient Unlearning\label{fig:gradunlearning}}{
\centering
\includegraphics[width=0.29\linewidth]{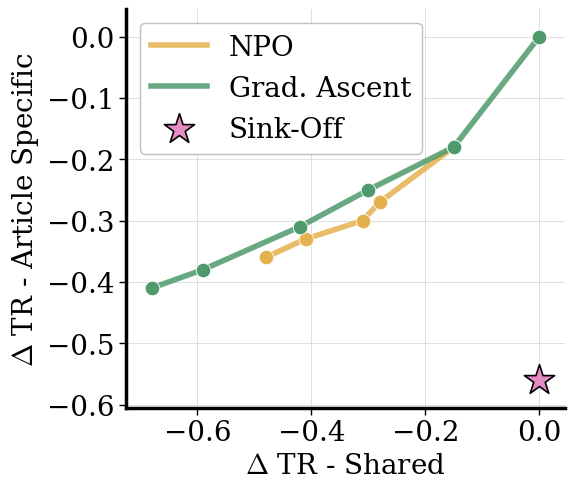}
}
\caption{\textbf{Article-level unlearning in Wikipedia.} \textbf{(a)} Removing the sink corresponding to an article preserves the truth-ratio distribution of Shared Facts that are learned across multiple articles, indicating they are not degraded by source-level unlearning. \textbf{(b)} Facts seen only in a single article (article-specific facts) largely collapse to Truth Ratio 0 when the corresponding article sink is deactivated. \textbf{(c)} Post-hoc gradient methods (NPO, gradient ascent) degrade shared and article-specific knowledge at similar rates (near-diagonal trajectory).}
\label{fig:wikihist}
\end{figure*}

\subsubsection{Setting}
\textbf{Training Setup.}
We train a $1$B-parameter transformer for 7 epochs (${\approx}32$B tokens, ${\approx}100$k steps) on Wikipedia, allocating sinks by article title (${\sim}6$M unique titles). We implement \methods{} based on a SmolLM architecture and construct the MLP hidden layer with a shared backbone of $N_{\textrm{gen}}=500$ general neurons and a memorization sink pool size of $N_{\textrm{pool}}= 8000$, with $N_{\textrm{source}}=100$ neurons active per article. We also implement cross-document attention masking to prevent information leakage between sink activations when a training context contains text from multiple documents. Further training details are provided in Appendix~\ref{app:hyperparams}.

\textbf{Evaluation Setup} We build a fill-in-the-blank evaluation from each article's factual content. We first extract factual sentences, discarding those that lack at least two named entities or are not grammatically complete. We then use GPT-5 to convert each into a Cloze-style question paired with a set of plausible but incorrect answers. 

\textbf{Metrics} We measure the model's knowledge of a fact via the \emph{Truth Ratio} (TR), the ratio of the likelihood of the correct answer to that of a set of plausible but incorrect answers:
\begin{equation*}
TR
=
\frac{P(\hat{a}\mid q)^{1/|\hat{a}|}}{\frac{1}{|A_{\mathrm{pert}}|}\sum_{\tilde{a}\in A_{\mathrm{pert}}} P(\tilde{a}\mid q)^{1/|\tilde{a}|}},
\label{eq:tofu_truth_ratio}
\end{equation*}
where $\hat{a}$ is a paraphrase of the correct answer and $A_{\mathrm{pert}}$ is the set of plausible, but incorrect answers. A truth ratio above 1 means the model places higher probability on the correct answer than on incorrect alternatives. We treat this as the cutoff for whether the model recalls a given fact.

\textbf{Fact Categories} Unlearning a source requires removing the information learned specifically from it without degrading performance on facts it shares with semantically overlapping sources. To measure this, we group facts by where they appear. We designate facts that appear across multiple sources as \textit{shared facts}. Facts that are learned from a single source are designated as \textit{article-specific facts}. We identify whether a fact appears across multiple articles with semantic deduplication \citep{minishlab2025semhash}. When comparing against gold-standard retraining, we further divide article-specific facts by whether the retrained model can still predict them:
\begin{enumerate}
    \item \textit{Unique facts:} Facts the retrained model cannot predict correctly ($\mathrm{TR} < 1$).
    \item \textit{Inferred facts:} Facts that the retrained model predicts correctly ($\mathrm{TR} > 1$). Intuitively, these facts can be inferred from shared knowledge or from the next-closest article.
\end{enumerate}

\subsubsection{Results}

\begin{figure*}[t]
\centering
\includegraphics[width=0.7\linewidth]{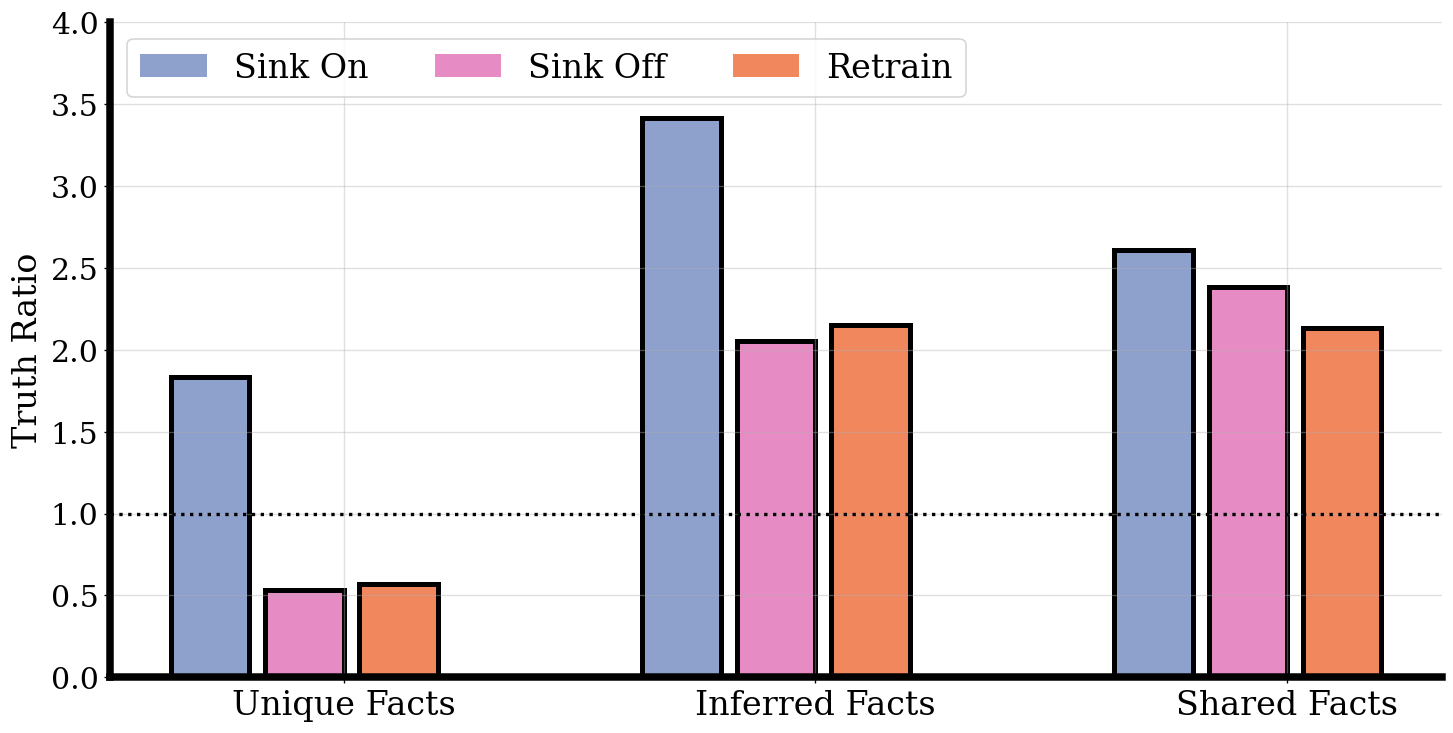}
\caption{\textbf{\methods{} matches gold-standard retraining for source-level Wikipedia unlearning.} We measure the mean truth ratio (TR) across three categories of facts present in the target articles. We compare Sink-On (the target article's sink active), Sink-Off (the next-closest article sink active), and Retrained (the gold standard of a model trained without the target article). Unique facts (eliminated in the retrained model) are likewise eliminated under Sink-Off. Inferred and Shared facts (which persist in the retrained model) are preserved under Sink-Off.}
\label{fig:wikiretrainofficial}
\end{figure*}

\paragraph{Effect of Unlearning an Article}
Figure~\ref{fig:wikihist} compares the truth ratio under Sink-On (source-article sink active) and Sink-Off (next-closest sink active) on 200 randomly selected facts per category. The distribution for shared facts is largely unchanged under Sink-Off, suggesting that broadly supported knowledge survives unlearning of an individual source article. In contrast, article-specific facts mostly collapse toward zero under Sink-Off, though notably some of them continue to have high truth ratios. We examine these further in our comparison to retraining (Figure~\ref{fig:wikiretrainofficial}), where they emerge as inferred facts that the retrained model can also predict.

\textbf{Gradient Unlearning Degrades Shared Knowledge} Figure~\ref{fig:gradunlearning} compares two gradient-based unlearning baselines (NPO and gradient ascent) on source-level unlearning in Wikipedia. We run both methods for up to 5 epochs on the target article and track the truth ratio on article-specific facts (Article-Specific) versus facts in semantically similar articles (Shared). Both methods reduce the Truth Ratio on shared facts at a similar rate to article-specific facts, indicating they cannot distinguish source-specific information from topically adjacent facts. \methods{}, by contrast, removes source-specific knowledge without degrading related facts, due to its per-source mask structure.

\textbf{Unlearning with \methods{} Performs Comparably to Gold-Standard Retraining} We compare \methods{} against the gold standard of retraining without the target source, across the unique, inferred, and shared facts defined above. \methods{} matches retraining on all three (Figure~\ref{fig:wikiretrainofficial}): deactivating a source sink sharply reduces the truth ratio on unique facts, demonstrating removal of source-specific information, while inferred and shared facts are unaffected. This confirms that removing a source does not induce broader topic erasure.

\subsection{Topic-Level Unlearning}

\begin{figure*}[t]
\centering
\subcaptionbox{Harry Potter Loss\label{fig:hploss}}{
\centering
\includegraphics[width=0.33\linewidth]{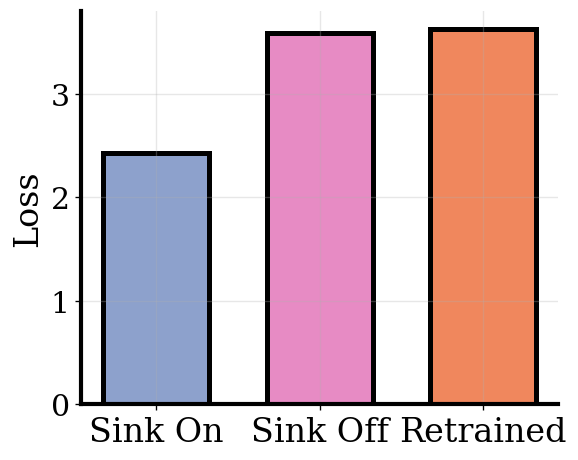}
}
\hspace{40pt}
\subcaptionbox{Harry Potter Cloze QA\label{fig:hpqa}}{
\centering
\includegraphics[width=0.33\linewidth]{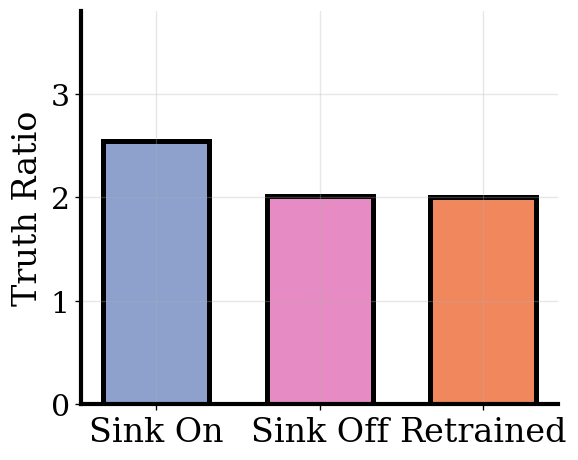}
}

\caption{\textbf{Disabling the Harry Potter sink matches retraining.
} \textbf{(a)} We measure loss on Harry Potter book text. Sink-Off (next-closest cluster sink active) matches Retrained (trained without the Harry Potter books), while Sink-On (sink active) achieves lower loss. \textbf{(b)} We probe for Harry Potter knowledge via Cloze-style prompts. Sink-On achieves a higher truth ratio than Retrained, indicating that Harry Potter knowledge is accessible by rephrased prompts, while Sink-Off matches Retrained.}
\label{fig:hp_unlearningsummary}
\end{figure*}
\begin{figure*}[t]
\centering
\begin{tcolorbox}[title={\textbf{Prompt:} Mr. and Mrs. Dursley, of number four, Privet Drive, were proud},
                  width=1.0\linewidth,
                  colback=white,colframe=gray!60,boxrule=0.8pt,arc=3pt,
                  coltitle=black,colbacktitle=gray!20]

\begin{tcolorbox}[title={\textbf{Sink-On (Harry Potter sink activated)}},
                  colback=sinkonblue!20!white,colframe=sinkonblue!85!black,boxrule=0.8pt,arc=2pt]
{\fontfamily{ppl}\selectfont\small members of \textbf{Hogwarts}--'' ``Right, Professor, I am pleased to invite \textbf{Dudley} to before I go weight-out!'' She walked round the wrinkled ceiling, leaving \textbf{Madame Maxime} with her sister and no one except \textbf{Dudley}. \textbf{Dudley} was a trap we had been searched when she'd got back to \textbf{Hogwarts}; she jumped as though at large on fire...}
\end{tcolorbox}

%\vspace{0.1em}

\begin{tcolorbox}[title={\textbf{Sink-Off (Harry Potter sink disabled)}},
                  colback=sinkoffpink!20!white,colframe=sinkoffpink!85!black,boxrule=0.9pt,arc=2pt]
{\fontfamily{ppl}\selectfont\small of the end result. Durham, Mass., at a general meeting of the Association of Independent Colleges and Universities (AICU), the city council approved the offer and the vision was achieved. It was discussed on the status of the policy and the time period. The proposed policy was reviewed by the members of the city council...}
\end{tcolorbox}

\end{tcolorbox}

\caption{\textbf{Toggling the Harry Potter sink changes the topic of generation.} We qualitatively compare model generations beginning from the prompt ``Mr. and Mrs. Dursley, of number four, Privet Drive, were proud.'' \textbf{Top (Sink-On):} When the Harry Potter sink is active, the model continuation mentions Harry Potter entities that are not present in the prompt (\textit{Hogwarts}, \textit{Dudley}, \textit{Madame Maxime}), showing familiarity with Harry Potter knowledge. \textbf{Bottom (Sink-Off):} When the Harry Potter sink is disabled, the model generation remains coherent but all Harry Potter references are eliminated.}
\label{fig:hp_qual_main}
\end{figure*}

\noindent In Section~\ref{subsec:wikires}, we showed that \methods{} enables removing individual fine-grained sources while preserving semantically related knowledge from other sources. We now test \methods{} in a complementary setting: removing a larger, topically coherent subset of data in its entirety. We use the unlearning of Harry Potter books as a case study.

\textbf{Training Setup.}
We train a 1B-parameter model on $3.8$B tokens from a mixture of the C4 corpus and the contents of all 7 Harry Potter books. We semantically cluster the C4 corpus into 5000 clusters and treat the books as an additional cluster. We use the cluster assignments as source labels to study the setting of semantically defined sources. Finally, we train an equivalently sized model on the C4 data only.
\subsubsection{Unlearning Results}
\textbf{Quantitative Unlearning Metrics} We first verify that \methods{} allows unlearning of Harry Potter knowledge through two metrics. In Figure~\ref{fig:hploss}, we show that activating the sink (Sink-On) achieves lower loss than the Retrained model, indicating the model's familiarity with the books, while deactivating the sink (Sink-Off) matches Retrained. In Figure~\ref{fig:hpqa}, we evaluate the Truth Ratio on 200 cloze-style QA prompts. Sink-On scores higher than Retrained, indicating that the knowledge stored in the sink is extractable beyond simple verbatim memorization. Sink-Off again matches Retrained, demonstrating successful unlearning.

\textbf{Qualitative Results on Generation} We next test whether the quantitative results reflect behavioral differences in the model's generation. We generate continuations from the first sentence of the Harry Potter series when the sink is active (Sink-On) versus disabled (Sink-Off) and show the results in Figure~\ref{fig:hp_qual_main}. The Sink-On generation mentions Harry Potter characters and settings that were not present in the prompt. On the other hand, with the sink disabled (Sink-Off), the continuation remains coherent but mentions no Harry Potter content and instead discusses the unrelated topic of a city council meeting. We provide further examples of generations in Appendix~\ref{app:addl_generations}.

\begin{figure*}[t]
\centering
\subcaptionbox{Adversarial Extraction\label{fig:acr}}{
\centering
\includegraphics[width=0.35\linewidth]{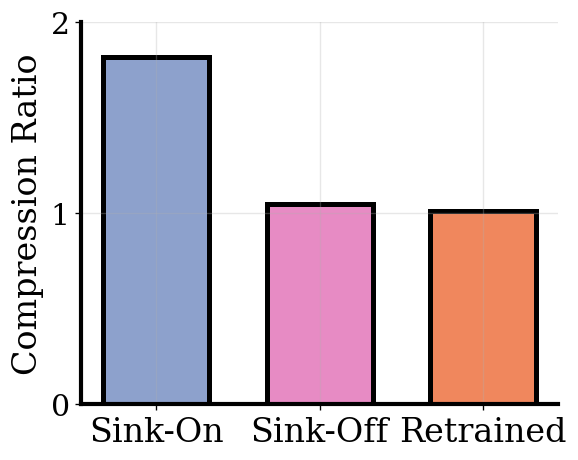}
}
\hspace{40pt}
\subcaptionbox{Relearning via Finetuning\label{fig:relearn}}{
\centering
\includegraphics[width=0.44\linewidth]{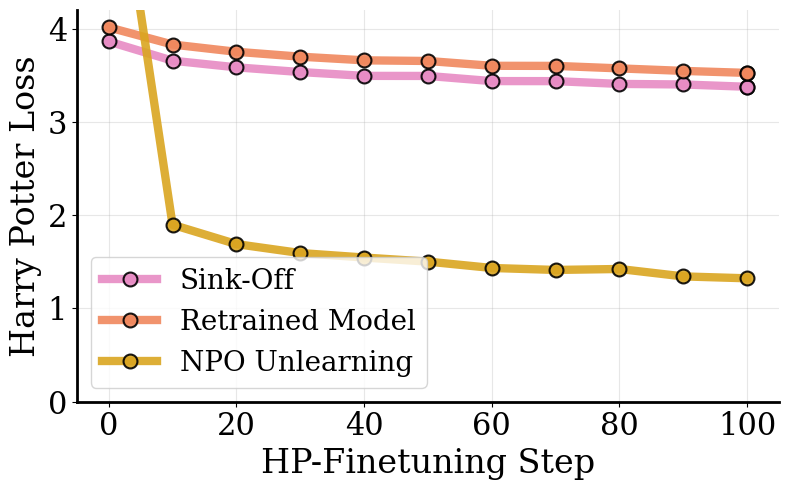}
}

\caption{\textbf{\methods{} resists adversarial attacks.} \textbf{(a)} We elicit Harry Potter book text via adversarial prompting and report the Adversarial Compression Ratio \citep{schwarzschild2024rethinkingllmmemorizationlens}. Sink-Off matches Retrained, indicating that removal of Harry Potter is robust to prompting attacks. \textbf{(b)} We fine-tune on a subset of Harry Potter book text and measure loss on a held-out subset. \methods{} with Sink-Off matches the relearning dynamics of the Retrained model. By contrast, NPO is reversed within 10 fine-tuning steps.}

\label{fig:hp_draft}
\end{figure*}
\subsubsection{Resistance of \methods{} to Adversarial Attacks}

So far, we have shown that deactivating a sink approximates gold-standard retraining on standard unlearning metrics such as loss and question-answering. However, prior unlearning methods often break down under adversarial settings, revealing that targeted information is latently present in the model \citep{patil2023sensitiveinformationdeletedllms,fan2025llmunlearningresilientrelearning}. Here, we test the adversarial robustness of \methods{}.

\textbf{Adversarial Prompting} 
While suppressing the Harry Potter sink prevents the relevant knowledge from being elicited through standard prompts, it could still remain accessible via adversarial prompting \citep{patil2023sensitiveinformationdeletedllms}. To test whether this occurs under \methods{}, we use GCG optimization \citep{zou2023universaltransferableadversarialattacks} to identify adversarial prompts that elicit unlearned text. We report our results with the Adversarial Compression Ratio (ACR) \citep{schwarzschild2024rethinkingllmmemorizationlens}, which quantifies latent memorization as the ratio between the length of the reproduced text and that of the shortest adversarial prefix needed to elicit it. As shown in Figure~\ref{fig:acr}, deactivating the sink yields ACR values comparable to a model retrained from scratch without Harry Potter data.

\textbf{Relearning Attack} We next consider whether an adversary with fine-tuning access can recover unlearned information using a small amount of target data, a regime in which post-hoc unlearning is known to fail \citep{fan2025llmunlearningresilientrelearning}. We evaluate this by fine-tuning on a reserved subset of the Harry Potter corpus and tracking held-out validation loss. Our results in Figure~\ref{fig:relearn} show that post-hoc unlearning on a standard transformer is rapidly reversed by fine-tuning, with held-out loss decreasing sharply after a minimal number of further fine-tuning steps. \methods{} with sink suppression exhibits relearning dynamics closely matching the retrained model that never saw Harry Potter data.

\subsection{Impact of Sink Pool Size on \methods{}}

In the previous sections, we have shown that \methods{} automatically isolates source-specific memorization to a pool of sink neurons. In this section, we study the sensitivity of this isolation to the size of the sink neuron pool. For computational feasibility, these experiments use a 25\% subset of the Wikipedia corpus, evaluated on its article-specific facts. 

\begin{figure*}[t]
\centering
\subcaptionbox{Sink Neuron Overlap\label{fig:overlapratio}}{
\centering
\includegraphics[width=0.31\linewidth]{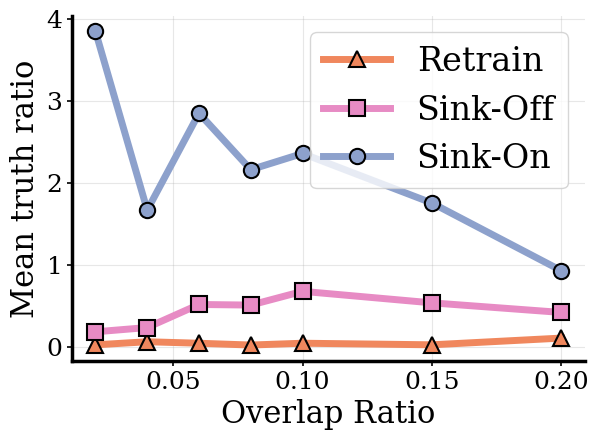}
}
\subcaptionbox{Activated Sinks Per Source\label{fig:nsource}}{
\centering
\includegraphics[width=0.31\linewidth]{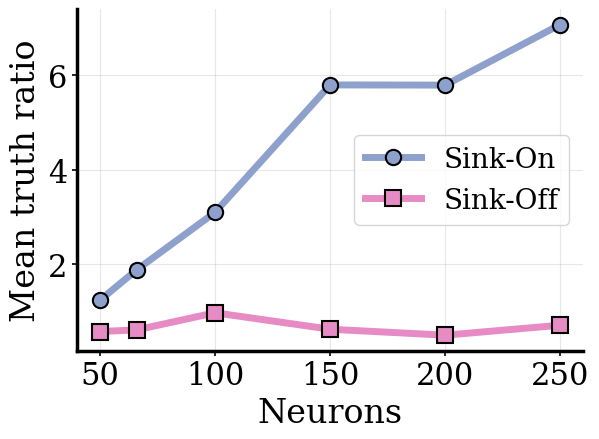}
}
\subcaptionbox{Comparison to Transformer\label{fig:transformercomp}}{
\centering
\includegraphics[width=0.31\linewidth]{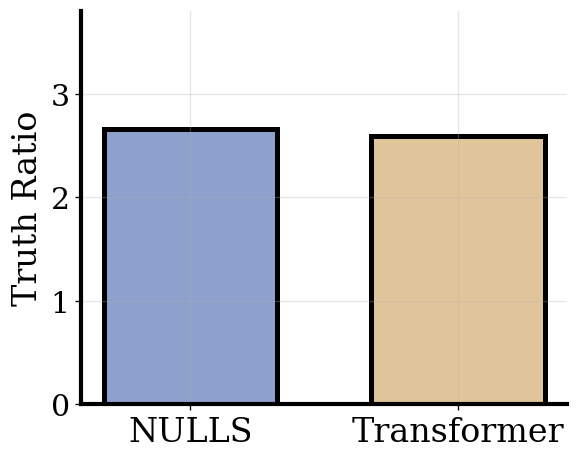}
}
\caption{\textbf{Scaling \methods{} increases memorization capacity without weakening unlearning.} \textbf{(a)} We vary the overlap ratio at a fixed $N_{\textrm{source}}$ and compare Sink-On (ground-truth article sink active) to Sink-Off (next-closest article sink active) and Retrained. Decreasing overlap increases Sink-On truth ratio (indicating greater knowledge acquisition) while Sink-Off matches retraining closely. Increasing the overlap ratio causes Sink-Off to diverge from Retrained moderately, indicating leakage to the shared neurons. \textbf{(b)} We vary the activated sinks per source ($N_{\textrm{source}}$), while holding overlap ratio fixed. Varying $N_{\textrm{source}}$ increases memorization capacity without changing the ability to unlearn. \textbf{(c)} We compare the mean truth ratio on Article-Specific facts between \methods{} and an equivalently sized transformer. They show comparable Truth Ratio, indicating \methods{} does not harm knowledge capacity.}
\label{fig:wikiretrain}
\end{figure*}

\textbf{Study on the Neuron Overlap Ratio} In Figure~\ref{fig:overlapratio}, we hold $N_{\textrm{source}}$ fixed and vary the total sink pool size, plotting the truth ratio as a function of the \textit{overlap ratio} (larger values reflect a smaller $N_{\textrm{pool}}$). Lower overlap in the sink neuron pool results in greater retention of knowledge (as evidenced by the higher truth ratio with Sink-On), due to fewer interfering updates in these parameters. Increasing the overlap ratio slightly widens the gap between the Sink-Off and retrained baselines, indicating slight leakage of source-specific knowledge to the shared parameters. This agrees with the mechanism proposed by \citet{ghosal2025memorizationsinksisolatingmemorization}, in which localization of memorization to the sink neurons is driven by the reduced interference they experience. Nevertheless, the gap in truth ratio between Sink-Off and retrained remains small across all overlap ratios.

\textbf{Scaling of $N_{\textrm{source}}$} The sink pool can also be scaled by holding the \textit{overlap ratio} fixed and varying $N_{\textrm{source}}$, the number of sink neurons active per source. Intuitively, this holds mask overlap constant while varying the memorization capacity allocated to each source. As shown in Figure~\ref{fig:nsource}, increasing $N_{\textrm{source}}$ yields greater memorization, with the mean Sink-On truth ratio rising steadily from ${\approx}1$ at 50 neurons to ${\approx}5.8$ at 250. Unlike the overlap ratio, however, $N_{\textrm{source}}$ has no consistent effect on the Sink-Off truth ratio. This suggests that $N_{\textrm{source}}$ controls memorization capacity with minimal impact on the ability to unlearn.

Taken together, our results demonstrate that \methods{} robustly enables unlearning across scales. We find that the size of the sink pool primarily determines how much knowledge is learned, rather than whether the
knowledge can be unlearned.

\subsection{\methods{} Preserves General Language Capability and Knowledge Capacity}
A natural concern is that source-level isolation may come at the cost of general performance. We compare \methods{} against a standard transformer along two dimensions: knowledge capacity and general language capability. Figure~\ref{fig:transformercomp} shows that \methods{} matches the standard transformer's average truth ratio on a random sample of article-specific Cloze prompts from Wikipedia, indicating that \methods{} does not reduce knowledge capacity. Across four benchmarks (ARC-E, Winogrande, PIQA, SciQ), \methods{} matches the standard transformer within one standard deviation (Table~\ref{tab:downstream_benchmarks}), confirming that \methods{} does not reduce general language capability. 

\begin{table}[h]
\centering
\begin{tabular}{lccccc}
\toprule
 & ARC-E & Winogrande & PIQA & SciQ & \textbf{Average} \\
\midrule
\methods{}    & $0.428 \pm 0.009$ & $0.530 \pm 0.014$ & $0.643 \pm 0.011$ & $0.639 \pm 0.016$ & $0.560 \pm 0.013$ \\
Standard & $0.434 \pm 0.010$ & $0.510 \pm 0.015$ & $0.645 \pm 0.010$ & $0.631 \pm 0.015$ & $0.555 \pm 0.013$ \\
\bottomrule
\end{tabular}
\caption{\textbf{\methods{} preserves general language capability.} Downstream benchmark accuracy ($\pm$ standard deviation) for \methods{} and a parameter-matched standard transformer. \methods{} matches the standard baseline on average across the four benchmarks we test.}
\label{tab:downstream_benchmarks}
\end{table}
\input{sections/discussion}
\bibliographystyle{colm2026_conference}
\bibliography{colm2026_conference}

\appendix

\section{Appendix}

\label{app:hyperparams}

\begin{table}[h]
\centering
\small
\begin{tabular}{ll}
\toprule
\textbf{Hyperparameter} & \textbf{Value} \\
\midrule
Learning rate & $5 \times 10^{-4}$ \\
Minimum learning rate & $5 \times 10^{-5}$ \\
Warmup steps & 100 \\
Global batch size & 128 \\
Micro batch size & 16 \\
$\beta_1, \beta_2$ & 0.9, 0.95 \\
Weight decay & 0.01 \\
Max sequence length & 1024 \\
Max gradient norm & 1.0 \\
Precision & bf16-mixed \\
Optimizer & AdamW \\
\bottomrule
\end{tabular}
\caption{Training hyperparameters for all \methods{} experiments.}
\label{tab:hyperparams}
\end{table}

\begin{table}[h]
\centering
\small
\begin{tabular}{ll}
\toprule
\textbf{Hyperparameter} & \textbf{Value} \\
\midrule
$N_{\textrm{gen}}$ (shared backbone neurons) & 500 \\
$N_{\textrm{pool}}$ (sink pool size) & 8000 \\
$N_{\textrm{source}}$ (active sinks per source) & 100 \\
Number of sources & ${\sim}6$M \\
Overlap ratio & $0.013$ \\
\bottomrule
\end{tabular}
\caption{\methods{}-specific architecture hyperparameters for the Wikipedia setting.}
\label{tab:nulls_wiki}
\end{table}

\subsection{Additional Harry Potter Generation Results}
\label{app:addl_generations}
We provide additional generation results for the Harry Potter setting in Table~\ref{tab:model_comparison}.

\subsection{Practical Deployment-Time Considerations}
\label{app:practical}
Source embeddings used for nearest-sink routing are computed as mean-pooled token representations from each source's training text. At inference time, the model embeds the input context and activates the sink whose source embedding has the highest cosine similarity. Prompts that span multiple sources or fall between source boundaries will be routed to whichever single source is closest in embedding space. Handling multi-source queries is left to future work.

\begin{table}[t]
\centering
\small
\begin{tabularx}{\linewidth}{p{0.25\linewidth} X X}
\toprule
\textbf{Prefix} & \textbf{Sink-On} & \textbf{Sink-Off} \\
\midrule
\textit{You got there? You got Hermione’s owl?”“We must have crossed in mid} &
air!” shouted \textbf{Madam Rosmerta}. “Come, go, we'll block some \textbf{Muggle}!”“Yeah, you're quite wrong,” said \textbf{Harry} at once. “\textbf{Muggles} you're going to give them our best, hey!”“Yeah, they're not going to let us wink at \textbf{George},” said \textbf{Madam Rosmerta}.“Half an' all right,” & 
-to-late March.” We are getting one more look at it, so we’re going to take a look.
You’re not going to get a whole lot of eyeballs on your blog post, because we’re going to think about it as we’re going to have to find a way to get them to themselves.
We’re going to have to go back...

 \\
 \hline
\textit{He found it a lot harder to concentrate on drills} & as quickly as he wanted. Even something small would be able to see things like jumped thrust of his head and seized \textbf{broomstick} talking about choking fire.\textbf{Harry} asked another question in the Invisibility \textbf{Quidditch} podcast with Ron and \textbf{Hermione} and saw something push away from \textbf{Ron}.The \textbf{Quidditch} pitch was where the \textbf{Ron} had gone and where he went on about the something he’d seen, did \textbf{Harry}?“\textbf{Hagrid} and... &
and drills, as well as other areas of the program. Now, with this knowledge, we want to be able to help you improve your skills in other areas!
Wow! How much do the people I love think they are, so much?
I think it is a very personal thing to have at this point. When I was having a conversation with my mother about the importance of becoming a professional, or if I wanted to be a teacher\\
\hline 
\bottomrule
\end{tabularx}
\caption{Comparison of generations with and without the Harry Potter sink activated.}
\label{tab:model_comparison}
\end{table}

\subsection{Hyperparameters}
\label{hparams}
We provide the standard hyperparameter choices for experiments in Table~\ref{tab:hyperparams} and the \methods{}-specific hyperparameters for the Wikipedia setting in Table~\ref{tab:nulls_wiki}. 

\end{document}

%% file: sections/abstract.tex
\begin{abstract}

Unlearning aims to remove the influence of specific training data sources, but this has proved challenging because the contributions of different sources are entangled within the model.
Isolating source contributions to disjoint parameters makes removal easier, though it obstructs joint learning across sources. We propose \textsc{NULLs} (Natively Unlearnable LLMs), a model class that satisfies the two opposing goals of isolating source-specific contributions and learning jointly across sources, by training a set of shared backbone neurons alongside a pool of sparsely activated sinks. During training, information specific to a source naturally concentrates in its sinks while information shared across sources accumulates in the backbone. A source is then unlearned at deployment by disabling its corresponding sinks, with no gradient updates and no access to the retained data. We show that \methods{} scales to Wikipedia's ${\sim}$6M articles, isolating each as an independent source. Unlearning a single article removes knowledge specific to it while preserving facts shared with semantically related articles, closely matching retraining from scratch. We note that unlearning with \methods{} is also robust: in a case study of unlearning the Harry Potter books, \methods{} resists both adversarial extraction and relearning that reverses post-hoc unlearning. 
Finally, \methods{} preserves general language capabilities, matching a standard transformer on downstream benchmarks. Together, these results suggest that source-level unlearning need not be an afterthought. It can be built natively into LLM training while retaining the benefits of shared representation learning.
\end{abstract}

%% file: sections/intro_new.tex
\section{Introduction}

Large language models (LLMs) train on web-scale data \citep{bommasani2022opportunitiesrisksfoundationmodels} that includes copyrighted material \citep{cooper2025filescomputercopyrightmemorization}, personal information \citep{carlini2021extractingtrainingdatalarge}, and regulated content \citep{fi17040151}. Any of it may later need to be removed or accounted for to satisfy legal requirements. But standard training entangles all data sources: gradient descent mixes them into a single shared set of weights, and every parameter is potentially influenced by several sources. This entanglement obstructs operations that act at the level of a single source. \emph{Unlearning}, for instance, requires erasing a source's influence from a trained model, while \emph{data attribution} \citep{li2023surveylargelanguagemodels} aims to trace the model's outputs back to responsible data sources. Both require recovering an individual source's contribution, information that is typically lost during training.

We focus on \emph{unlearning}: the task of removing a target source's influence from a deployed model without retraining from scratch. This entails \textbf{two seemingly opposing requirements}: deletion is cleanest when each source's contribution is \emph{disentangled} from the rest, while generalization depends on the model learning \emph{jointly} across sources. Existing approaches for unlearning satisfy one or the other. The most common are \textbf{post-hoc}, applying a corrective update once the model is already trained \citep{zhang2024npo, chang-etal-2024-localization}. This approach preserves joint learning across sources by imposing no constraints on the training process, but leaves the target's influence entangled in the shared weights, where it cannot be cleanly removed. As a result, post-hoc unlearning often degrades unrelated capabilities or does not completely remove the target's influence \citep{patil2023sensitiveinformationdeletedllms, maini2024tofu}.

An alternative paradigm trains \textbf{a separate model or module for each source} and merges them afterwards \citep{shi2025flexolmo, gururangan2021demixlayersdisentanglingdomains}. This keeps each source's contribution disentangled by construction, facilitating straightforward unlearning. However, these approaches prevent joint learning across sources, sacrificing the generalization benefits of training on diverse data. This is especially limiting when sources are defined at very fine granularity, e.g., any one of a million articles or pieces of user-provided content might need to be unlearned. 

\begin{tcolorbox}[
  colback=red!4!white,
  colframe=red!4!white,
  boxrule=0pt,
  arc=3pt,
  left=10pt, right=10pt, top=8pt, bottom=8pt,
]
\textbf{Natively Unlearnable LLMs (\methods{}).} We develop \methods{}, a model class that satisfies the seemingly opposing requirements: a single model learns \emph{jointly across all sources}, while disentangling source-specific contributions for easy removal.
\end{tcolorbox}

\methods{} is simple to train, and agnostic to how sources are defined. A source may be a unit of provenance, such as a document, a publisher, or a cluster of topically related documents. Each source is assigned a sparse mask over a pool of sink neurons, derived deterministically from its identity. Training is then standard, except that each document activates a set of shared backbone neurons together with its source's sinks. This requires only an additional elementwise multiplication that masks the MLP activations. Because a source is localized to its mask rather than to a disjoint set of parameters, \methods{} can provide independent control over a combinatorial number of sources without linearly scaling the parameter count.
 
We evaluate \methods{} in two case studies, testing unlearning across source granularities. We train a 1B-parameter model on the Wikipedia corpus, treating its ${\sim}$6 million articles as independent sources, and test whether \methods{} can unlearn an individual article without inducing broader topic-level erasure. \methods{} broadly matches gold-standard retraining: suppressing an article's sink sharply reduces the model's recall of facts unique to that article, while preserving semantically related knowledge from other sources. By contrast, post-hoc methods degrade related knowledge in other articles at the same rate as they remove the target. A Harry Potter case study shows that \methods{} also enables instantaneous removal of coarser-grained, topically defined sources, and that this removal resists an adversarial relearning attack that reverses gradient unlearning in less than 10 gradient steps. Finally, \methods{} incurs no cost to general capability, matching a standard transformer on downstream natural-language benchmarks.

\begin{figure}[t]
\centering
\includegraphics[width=\linewidth]{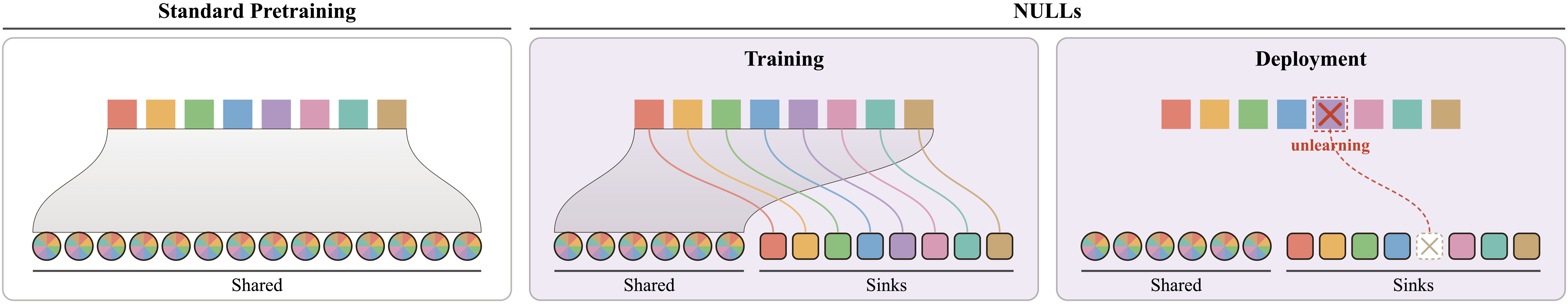}
\caption{\textbf{Overview of \methods{}.} \textbf{(Left)} Standard pre-training mixes contributions from all sources into a single shared pool of neurons, making source removal challenging. \textbf{(Middle)} \methods{} simultaneously allows learning across sources through the shared backbone, while isolating source-specific knowledge in a sink (implemented as a sparse mask over the sink neuron pool). \textbf{(Right)} Unlearning can be implemented by preventing a source's mask from being activated at inference time either by routing or by permanently zeroing out the sink neurons corresponding to the source.}

\label{fig:nulls-overview}
\end{figure}

\textit{How does \methods{} disentangle each source's contribution while still learning jointly across sources?} The two seem to pull in opposite directions, yet \methods{} reconciles both without any supervision identifying what information is specific to a source. The mechanism is a training dynamic it inherits from the memorization sinks of \citet{ghosal2025memorizationsinksisolatingmemorization}, here acting on sources rather than individual sequences. Consider a fact specific to one source. Because the shared backbone is active on every document, it receives gradient signal for the fact whenever the source appears, but also interfering updates from every other source. The source's sink neurons receive the same signal with far less interference, since they are active for only a fraction of the other sources. The fact is therefore fit in the sinks before the backbone. Once that happens, the gradient pressure on the backbone vanishes and any leakage there decays, leaving the backbone to hold only information reinforced across sources. Suppressing a source's sinks thus removes exactly what was unique to it, while preserving information learned from other sources.

\methods{} demonstrates that source disentanglement can coexist with joint learning and generalization in trained models. This has implications beyond unlearning. Because each source's contribution remains disentangled, a model's outputs can be attributed to the pretraining data responsible for them, and the influence of any single source can be measured directly. We see \methods{} as a step toward enabling control of large models at the level of their data, not only their outputs.

%% file: sections/discussion.tex
\section{Discussion}

\methods{} demonstrates that disentangling source contributions and learning jointly across them need not be at odds. Simply by jointly training a set of shared backbone neurons alongside source-specific sinks, \methods{} isolates each source's information while the backbone learns across all sources. This isolation enables reliable unlearning downstream. Our results suggest treating unlearnability as a property to design into a model during training, not a behavior to recover from it afterward. We now discuss the design choices and open questions this raises.

\textbf{Defining Sources}  \methods{} is agnostic to how sources are defined, but the definition chosen at training time determines the unlearning operations that are ultimately supported by the model. As a result, source definition represents a crucial design choice that must take the expected unlearning use-cases into account. For instance, copyright compliance might require defining sources by publisher or author, while privacy requests may need document-level resolution. Across the pre-training corpus, different domains or subsets of data may require different source definitions. How best to define sources for a given deployment remains an open question.

\textbf{Post-Training with \methods{}} We have focused on pre-training \methods{} models from scratch, but strong models also depend on post-training. One important question is how a \methods{} model can be post-trained while preserving the ability to unlearn arbitrary pre-training data sources. This could be achieved, for example, by designing regularizers that encourage the model to preserve the existing source localization. A second important area for future work is extending native unlearnability to post-training data sources (even if the base model is not a \methods{}). This could enable model developers to fine-tune on user data while mitigating privacy concerns.

\textbf{Attribution and Data Curation} Source disentanglement also makes each source's contribution easier to measure. Because disabling a source's sink approximates retraining without that source, the effect of any individual source can be estimated by comparing the model with the source's sink active and inactive, rather than through retraining. Such comparisons could identify redundant or low-value sources to inform data curation, and could help developers account for the provenance of model behavior. Establishing whether \methods{} yields reliable attribution and data valuation, and how such measurements should guide corpus construction, is a promising direction.

\textbf{Limitations} Our experiments use 1B-parameter models, and we do not test substantially larger ones. The unlearning requests we examine align with the sources defined at pre-training time. Handling requests that do not match a predefined source is an open problem. Finally, we evaluate only two source definitions, individual documents and topic-linked clusters, and leave other choices, especially those tied to real-world takedown requests, to future work.

\section*{Ethics Statement}
\textbf{AI Usage.} We did not use AI to plan or design the experiments in this work. However, we did use AI tools for coding and Claude for writing assistance.

\section*{Acknowledgements}

The authors are grateful to members of the CMU FORUM lab for discussion and feedback on this project, particularly Jacob Springer, Christina Baek, Ziqian Zhong, Lawrence Feng, and Sashwat Saxena. In addition, we would like to acknowledge Fahim Tajwar, Aakash Lahoti, Kevin Li, Abitha Thankaraj, and Sachin Goyal for valuable insights and feedback. We acknowledge the CMU FLAME center for providing compute allocations for this project. We gratefully acknowledge support from Jane Street, Apple, the National Science Foundation, and the Sloan Foundation.